%% file: main.tex
\documentclass[10pt,twocolumn,letterpaper]{article}

\usepackage[pagenumbers]{cvpr} %

\input{preamble}
\definecolor{cvprblue}{rgb}{0.21,0.49,0.74}
\usepackage[pagebackref,breaklinks,colorlinks,allcolors=cvprblue]{hyperref}

\usepackage{colortbl}
\usepackage{xcolor}

\usepackage{amssymb}
\usepackage{dsfont}
\usepackage{bbm}
\usepackage{graphicx}

\usepackage{cuted}
\usepackage{capt-of}

\definecolor{r}{RGB}{255,200,200}
\definecolor{y}{RGB}{255,255,200}
\definecolor{p}{RGB}{255,222,189}

\def\paperID{172} %
\def\confName{CVPR}
\def\confYear{2025}

\title{HDGS: Textured 2D Gaussian Splatting for Enhanced Scene Rendering}

\author{Yunzhou Song$^{1}$\thanks{equal contribution, $^\dagger$ corresponding author} \quad Heguang Lin$^{1*}$ \quad Jiahui Lei$^{1\dagger}$ \quad Lingjie Liu$^{1}$ \quad Kostas Daniilidis$^{1, 2}$\\
$^1$University of Pennsylvania \quad $^2$Archimedes, Athena RC\\
{\tt\small \{timsong, hglin, leijh, lingjie.liu, kostas\}@upenn.edu}
\vspace{-2em}
}

\begin{document}

\maketitle
\begin{strip}\centering
\vspace{-1em}
\includegraphics[width=0.99\linewidth]{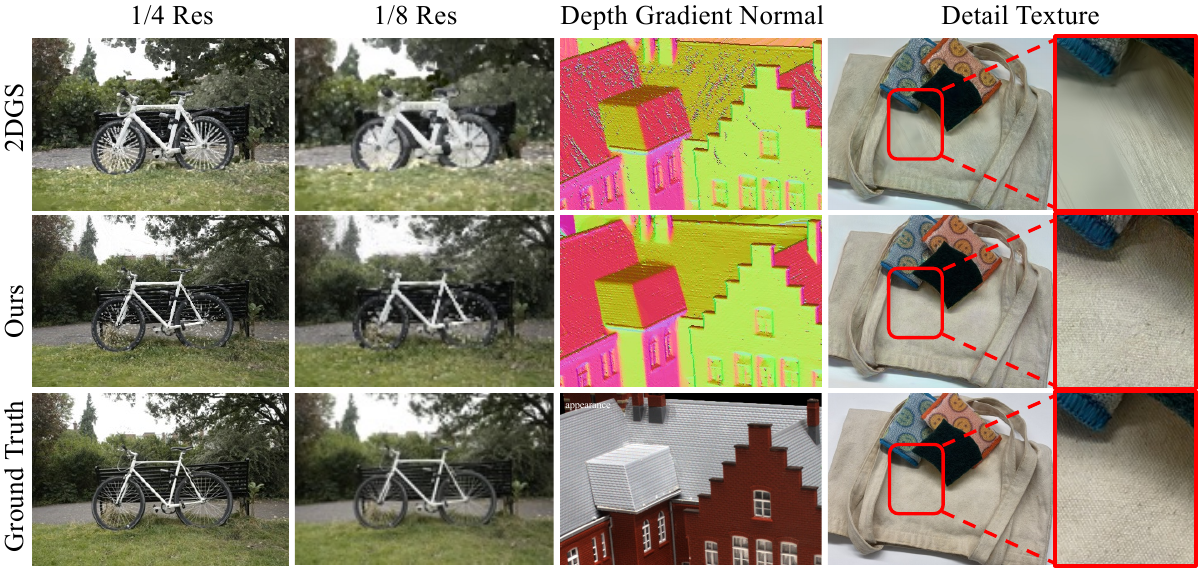}
\captionof{figure}{ Overview of the improvements in our method over 2D Gaussian Splatting (2DGS). \textit{First two columns}: Our frustum-based sampling technique effectively mitigates aliasing for high-frequency appearance rendering at reduced resolutions (1/4 and 1/8). \textit{Third column}: Our approach yields enhanced geometric consistency with smoother surface normals. \textit{Last column}: Our method disentangles geometry from appearance by utilizing a per-Gaussian texture map, allowing for rich detail preservation that 2DGS cannot resolve.}
\end{strip}

\input{sec/0_abstract}    
\input{sec/1_intro}

\input{sec/2_related}

\input{sec/3_method}

\input{sec/4_exp}

\input{sec/5_conclusion}

{
    \small
    \bibliographystyle{ieeenat_fullname}
    \bibliography{main}
}
\input{sec/X_suppl}

\end{document}

%% file: sec/0_abstract.tex
\begin{abstract}
Recent advancements in neural rendering, particularly 2D Gaussian Splatting (2DGS), have shown promising results for jointly reconstructing fine appearance and geometry by leveraging 2D Gaussian surfels. However, current methods face significant challenges when rendering at arbitrary viewpoints, such as anti-aliasing for down-sampled rendering, and texture detail preservation for high-resolution rendering. We proposed a novel method to align the 2D surfels with texture maps and augment it with per-ray depth sorting and fisher-based pruning for rendering consistency and efficiency. With correct order, per-surfel texture maps significantly improve the capabilities to capture fine details. Additionally, to render high-fidelity details in varying viewpoints, we designed a frustum-based sampling method to mitigate the aliasing artifacts.
Experimental results on benchmarks and our custom texture-rich dataset demonstrate that our method surpasses existing techniques, particularly in detail preservation and anti-aliasing.

\end{abstract}

%% file: sec/1_intro.tex
\vspace{-1em}
\section{Introduction}

\label{sec:intro}
\noindent Neural rendering has made remarkable strides in novel view synthesis and geometry reconstruction. Recently, 2D Gaussian Splatting (2DGS) ~\cite{Huang2DGS2024} emerged as a promising method that leverages Gaussian surfels to represent complex scenes. Despite its advantages, however, the current 2DGS technique encounters the challenge of rendering high-quality images from arbitrary resolution and distance. Close-up rendering often suffers from insufficient detail, whereas aliasing artifacts arise when rendering at low resolutions or from distant viewpoints. These flaws hinder the realism and quality of 3D rendering by compromising visual fidelity at varying resolutions and distances. 
This paper explores solutions to high-definition 3D scene reconstruction using 2D Gaussian Splatting, enabling visualization from arbitrary viewpoints and resolutions.

Gaussian Splatting struggles when appearance details exceed geometric resolution in close views. The entanglement of geometry and appearance in Gaussian Splatting requires numerous primitives for accurate representation. 
We proposed a novel method to assign each primitive with an optimizable texture map to disentangle the appearance with geometry. However, the straightforward implementation suffers from significant generalization issues and popping artifacts from novel views. We identify that the artifacts stem from 2DGS’s global per-view primitive sorting, which we address by switching to per-ray sorting ~\cite{radl2024stopthepop} as regularization. Additionally, since assigning a per-surfel texture map introduces higher memory demand, we apply Fisher-information-based pruning ~\cite{hanson2024pup} to remove Gaussians with high uncertainty to maintain an efficient representation.

On the other hand, the aliasing artifact has been an emerging issue in Gaussian Splatting when rendering from arbitrary viewpoints. Previous works ~\cite{yan2024multi,yu2024mip} proposed multiple filters to address this issue. However, adapting these techniques to 2DGS is non-trivial, as the projection of the 2D surfels onto the image plane is no longer Gaussian distributed. To address this issue in 2DGS, we propose a frustum-based sampling technique. By treating each pixel as a light frustum that samples over a volume, rather than a single ray from its center, our approach reduces aliasing artifacts in high-frequency regions and provides smoother rendering across varying resolutions.

Equipped with per-surfel texture maps, as well as per-ray sorting and pruning, our system can efficiently render high-quality images from close-up viewpoints. Augmented with a frustum-based sampling strategy, our pipeline significantly mitigates aliasing artifacts from distant viewpoints, resulting in an alias-free method to render high-quality images from arbitrary viewpoints. Our contributions are summarized as:
\begin{itemize}

    \item  We proposed HDGS, a 2DGS-based method with per-surfel texture maps that can be rendered correctly with per-ray sorting and maintain efficiency through Fisher-information-based pruning.

    \item We introduced a frustum-based Gaussian sampling method to mitigate high-frequency rendering aliases.
    
    \item Our pipeline achieves state-of-the-art rendering quality on standard benchmarks. We collected a texture-rich dataset, on which our method can capture fine details that other baseline methods fail to preserve. 
\end{itemize}

%% file: sec/2_related.tex
\section{Related Work}
\label{sec:relate}

\textbf{Anti-aliasing in novel view synthesis}. Following the pioneering neural radiance field (NeRF) \cite{mildenhall2020nerf}, numerous works \cite{barron2021mip,barron2022mip,hu2023tri} have introduced novel methods for anti-aliasing in neural rendering. Zip-NeRF \cite{barron2023zip} presented a multi-sampling strategy to enable anti-aliasing within a more efficient hash code-based representation \cite{muller2022instant} for neural radiance fields. Recently, Rip-NeRF \cite{liu2024rip} proposed a Ripmap-encoded Platonic solid representation for anti-aliased neural radiance fields. With the emergence of 3D Gaussian Splatting ~\cite{kerbl3Dgaussians} and its extensions in material modeling ~\cite{shi2023gir,jiang2024gaussianshader}, avatar creation~\cite{qian2024gaussianavatars,zielonka2023drivable}, and dynamic scene reconstruction ~\cite{yan2024street}, novel approaches have been developed to address aliasing in 3D Gaussian Splatting. For example, ~\cite{song2024sa, li2024mipmapgsletgaussiansdeform} applied scale-adaptive modifications during optimization, which can be used as a plugin in any pretrained 3D Gaussian splatting. ~\cite{yu2024mip, yan2024multi} analyze the approximated Gaussian primitive covariance after the affine projection and add post-process filters to suppress the high-frequency signals in 3D Gaussian Splatting. However, the 2D Gaussian Splatting adopts precise projection for each primitive onto the image plane, which is no longer Gaussian distributed. Therefore, we proposed a frustum-based sampling strategy to increase the sampling frequency to mitigate the anti-aliasing issues in 2DGS.

\noindent \textbf{Texturing 3D representations}. 3D representations can have different forms of parametrization, such as neural networks \cite{mildenhall2020nerf,sitzmann2020implicit,tancik2020fourier}, voxel grid \cite{fridovich2022plenoxels,newcombe2011kinectfusion,sun2022direct}, and a hybrid of feature grids and networks \cite{chen2022tensorf,liu2020neural,lombardi2021mixture}. Follow-up works texture these representations by optimizing a surface parametrization~\cite{das2022learning,srinivasan2025nuvo}, while other works disentangle appearance and geometry with combined methods~\cite{yang2022neumesh,huang2023nerf}. The most relevant two works have attempted to disentangle geometry and appearance by introducing texture maps in Gaussian Splatting ~\cite{rong2024gstex,xu2025texture}. The approach in~\cite{xu2025texture} applies the texture map with the assumption that Gaussians in the scene lie on a surface topologically similar to a sphere. Such an assumption failed in the scenes with complicated geometry. In~\cite{rong2024gstex}, the surfel nature of 2DGS is leveraged to train a per-Gaussian texture map after geometry reconstruction. However, this approach substantially increases computational demands, leading to failure on complicated scenes and not addressing aliasing artifacts.

\noindent\textbf{Popping Artifact of Rasterization}
Gaussian Splatting rasterizer sorts the center depth of all primitives once for each view and traverses them by the same global order along the ray. As a result, for the same 3D area the blending order of primitives may differ from training views and test views, which introduces the popping artifact. Methods including~\cite{radl2024stopthepop, kbuffer, sort, sort2} tried to approximately per-ray sort the primitives with limited resources and computing. While some other techniques design order-free blending methods like ~\cite{hou2024sortfreegaussiansplattingweighted}.

%% file: sec/3_method.tex
\begin{figure*}[t!]
    \centering
    \includegraphics[width=0.8\linewidth]{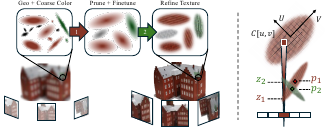}
    \caption{Our proposed method for high-quality scene reconstruction leverages the explicit per-ray depth and 2D ray-surfel intersection coordinates of 2D Gaussian Splatting. The left image illustrates our coarse-to-fine pipeline, which begins with 2DGS training, proceeds through Fisher pruning, and ends with per-surfel texture mapping for enhanced visual fidelity. The right image highlights our design of per-ray sorting strategy with primitive texture: though the center of surfel $p_2$ is closer to the screen than $p_1$, our method sorts the Gaussians correctly as $z_1$ and $z_2$ in the depth order.}
    \label{fig:pipeline}
    \vspace{-1em}
\end{figure*}

\section{Method}
\label{sec:method}

\subsection{Preliminaries}
\textbf{2D Gaussian Splatting.}
Huang at al.~\cite{Huang2DGS2024} proposed to represent 3D scenes with 2D Gaussian surfels to reconstruct the appearance and fine-level geometry jointly. Each Gaussian surfel is parameterized with two principal tangential axis vectors $\mathbf{t_u}$ and $\mathbf{t_v}$ paired with two scaling factors $s_u$ and $s_v$, the surfel center $\mathbf{p}$, opacity $\alpha$, and the spherical harmonics parameters $\mathbf{c}$. The surfel normal $\mathbf{n}$ is then defined by $\pm( \mathbf{t_u} \times \mathbf{t_v})$, where the direction is towards the viewing direction.

Given a pixel $\mathbf{x}$ in the image space, instead of projecting the primitive center to the image plane and approximating the projected covariance as the 3DGS does, the 2D Gaussian Splatting computes the explicit precise ray-surfel intersection $(u, v)$ in the surfel coordinates by the projection matrix $\mathbf{P}$ and the splat-to-world transformation $\mathbf{H}$ as
\begin{equation}
\mathbf{x} = (xz, yz, z, z)^\top  = \mathbf{P}\mathbf{H}(u, v, 1, 1)^\top, \quad
\end{equation}
the transformation $\mathbf{H}$ is defined by the surfel parameters as
\begin{equation}
    \mathbf{H} = 
\begin{bmatrix}
 s_u\mathbf{t_u} & s_v\mathbf{t_v}  & 0  & \mathbf{p}\\
 0 & 0 & 0  & 1
\end{bmatrix}
\end{equation}
Then the surfel weight is computed by querying a normalized Gaussian distribution as $G(\mathbf{x}) = \exp(-\frac{u^2+v^2}{2})$. The pixel color $\mathbf{C}$, depth $D$, and surface normal $\mathbf{N}$ are then accumulated across all intersected surfels along the ray from front to back by alpha-blending:
\vspace{-3mm}
\begin{equation}
     \mathbf{C(x)} = \sum_i \mathbf{c}_i\mathbf{(d)} \alpha_i G_i(\mathbf{x}) T_i
\end{equation}
\begin{equation}
    D(\mathbf{x}) =  \sum_i z_i(\mathbf{x}) \alpha_i G_i(\mathbf{x}) T_i
\end{equation}
\begin{equation}
    \mathbf{N(x)} =  \sum_i \mathbf{n}_i \alpha_i G_i(\mathbf{x}) T_i
\end{equation}
\vspace{-3mm}

 where $T_i = \prod_j^{i-1}(1-\alpha_j G_j)$ is the transmittance, and $\mathbf{d}$ is the viewing direction. During optimization, the 2DGS introduces two geometric regularizations Eq.\ref{eq:Ld}, Eq.\ref{eq:Ln} to capture fine-grained surfaces, where $\mathbf{N}_d$ is the normal map computed by the gradient of the depth map.
\begin{equation}
\label{eq:Ld}
     L_d = \sum_{i, j} \alpha_i G_i T_i \alpha_j G_j T_j (z_i-z_j)^2
\end{equation}
 \vspace{-1em}
\begin{equation}
\label{eq:Ln}
     L_n = \sum_i \alpha_i G_i T_i  (1 - \mathbf{n}_i \mathbf{N}_d)
\end{equation}

\noindent\textbf{Fisher Information for 3D Gaussian Pruning.}
Prior works have explored various approaches to compress 3D Gaussian-based scene representations, such as uncertainty quantification ~\cite{hanson2024pup,jiang2023fisherrf}, and hybrid method \cite{fan2023lightgaussian}. Fisher information provides a principled way to quantify the uncertainty in parameter estimates. In 3D scene reconstruction, this uncertainty arises from the inherently underconstrained nature of projecting 3D scenes onto 2D images. Gaussians not well-constrained by multiple camera views may be able to reconstruct the input views from a range of locations and scales. To quantify this uncertainty mathematically, we start with the $L_2$ error over the input reconstruction images $\textbf{I}_G$:

\vspace{-3mm}
\begin{equation}
\label{eq:l2loss}
L_2 = \frac{1}{2}\sum_{\phi \in P_{gt}} ||\textbf{I}_{\mathcal{G}}(\phi) - \textbf{I}_{gt}||^2_2
\end{equation}

The Hessian of this loss function captures the sensitivity of the error to changes in Gaussian parameters:

\vspace{-5mm}
\begin{align}
\label{eq:Hessien}
\nabla^2_\mathcal{G} L_2 = & \sum_{\phi \in P_{gt}} \nabla_\mathcal{G} \textbf{I}_\mathcal{G}(\phi)\nabla_\mathcal{G} \textbf{I}_\mathcal{G}(\phi)^T \nonumber \\
                  & + (\textbf{I}_\mathcal{G}(\phi) - \textbf{I}_{gt})\nabla^2_\mathcal{G} \textbf{I}_\mathcal{G}(\phi)
\end{align}
\vspace{-1em}

For a converged model where the reconstruction error approaches zero, the second term vanishes, leaving us with the Fisher approximation:

\vspace{-3mm}
\begin{equation}
\label{eq:Approx}
\nabla^2_\mathcal{G} L_2 =  \sum_{\phi \in P_{gt}} \nabla_\mathcal{G} \textbf{I}_\mathcal{G} \nabla_\mathcal{G} \textbf{I}^T_\mathcal{G}
\end{equation}

This approximation depends only on the input poses $P_{gt}$ and not the input images $I_{gt}$. We can then compute a per-Gaussian sensitivity score by taking the block Hessian for each Gaussian's parameters. Gaussians with lower sensitivity scores are candidates for pruning as they have less impact on the reconstruction error. 

\subsection{Frustum-based Sampling}

\begin{figure}
    \centering
    \hspace{-1em}
    \includegraphics[width=0.8\linewidth]{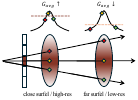}
    \caption{\textbf{Frustum Sampling.} The average density of the sampled primitive decreases as the surfel moves farther from the camera or as the sampling resolution decreases, effectively acting as a low-pass filter for anti-aliasing.}
    \label{fig:cone}
    \vspace{-2em}
\end{figure}

To model and render the high-frequency appearance details in diverse views and resolutions, we can not avoid tackling the anti-aliasing issues. The aliasing effect may occur when sampling a continuous signal by a discrete function especially when the sampling frequency is less than twice the signal frequency, known as the Nyquist Theorem. In the field of computer graphics and rendering, the sampling frequency is determined by the image's pixel resolution, while the signal frequency is related to the object's appearance complexity after projection to the image plane. Due to the feature of perspective projection, objects far from the camera with low appearance complexity may be crowded into small areas on the image plane, which leads to higher appearance frequency and may trigger aliasing.

Recent 3D Gaussian Splatting anti-aliasing techniques ~\cite{yu2024mip, yan2024multi} analyze the approximated Gaussian primitive covariance after the affine projection and add post-process filters to suppress the high-frequency signals. However, the 2D Gaussian Splatting adopts precise projection for each primitive, whose projection is no longer Gaussian distributed. As the post-process methods become non-trivial in the 2D Gaussian Splatting setting, we consider increasing the sampling rate to mitigate the aliasing effect by estimating the average Gaussian density. Each pixel on the sensor receives exactly the light of a frustum, rather than only one ray shooting out from the center. For each 2D pixel of size $(2\delta_x, 2\delta_y)$ sampling a surfel in 3D space, we then cast the center ray $\mathbf{x}$ and four corner rays to compute 5 intersections, and average the sampled Gaussian density as Eq.~\ref{eq:avg} for the following alpha-blending.

\vspace{-1em}
\begin{equation}
\label{eq:avg}
    \hat{G}(\mathbf{x}) = w_mG(\mathbf{x}) + \sum_{i, j \in \{1,-1\}} w_cG(\mathbf{x} + (i\delta_y, j\delta_x))
\end{equation}

We are not casting 5 individual subpixel rays for each pixel, which causes 5 times more computing, but compute the intersections of the frustum with the same surfel touched by the center ray. Please refer to supplementary materials for comparison with different sampling weights $w_m, w_c$. Illustrated in Fig.~\ref{fig:cone}, as the frustum diverges along distance, the high-frequency primitives in the distance will have a smaller weight, which equals low-pass filtering for anti-aliasing. Meanwhile, lower sampling resolution leads to a larger frustum diameter, which also suppresses the high-frequency details. Our method precisely and efficiently captures the high-frequency signals as the ray-surfel intersection is closed and has a low computing cost.

\begin{figure*}
\vspace{-1em}
    \centering
    \includegraphics[width=1.0\linewidth]{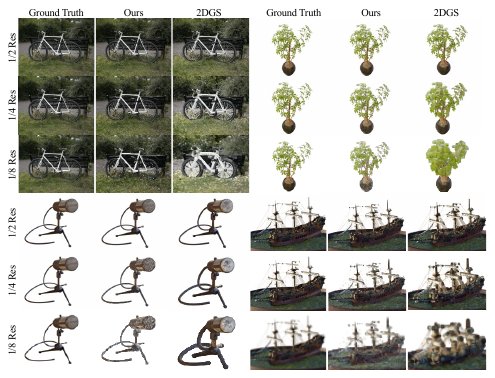}
    \caption{Comparison of our method and 2DGS at reduced resolutions (1/2, 1/4, 1/8) on NeRF synthetic dataset  \cite{mildenhall2020nerf} and  Mip-NeRF 360 dataset ~\cite{barron2022mip}. Our approach effectively mitigates dilation artifacts and aliasing problems in 2DGS. Note that the 1/8 rendering of the microphone by 2DGS has the same image scale as above, whose dilation aliasing is considerably severe.}
    \label{fig:alias}
    \vspace{-1em}
\end{figure*}

\input{tables/multires}

\subsection{Per-Ray Sorting with Fisher Pruning}
The Gaussian splatting rasterization sorts the primitives by their center depths per view before rendering and then traverses them in the same order for different pixel rays during alpha blending. As a result, popping artifacts occur due to the primitives' order difference of the same 3D position in different views, which hinders the novel view rendering fidelity. This phenomenon is more obvious when rendering high-frequency details or incorporating thin and large primitives. We leverage the explicit depth $z$ of 2D Gaussian Splatting to implement the per-ray k-buffer sorting ~\cite{kbuffer} and per-tile sorting in ~\cite{radl2024stopthepop}. We set the tile size as 8 and compute the center pixel depth of the tile to leverage the per-tile radix sorting. We set the ray sort buffer size k as 24 for synthetic scenes and 16 for real scenes for efficiency. 

While ray sorting significantly mitigates popping artifacts, it introduces computational overhead and can impact convergence due to increased ordering complexity between adjacent rays. Therefore, it becomes beneficial to reduce the number of surfels while preserving the scene's visual quality. To address these challenges, we introduce a Fisher information-based pruning strategy inspired by \cite{hanson2024pup} for our 2D surfel representation.

Given a 2D Gaussian surfel $G_i$, we compute its sensitivity score using a block-wise Fisher information matrix similar to equation \ref{eq:Approx}. As suggested in \cite{hanson2024pup}, we only consider the spatial position $x_i$ and scaling $s_i$ parameters when computing the Jacobian, as they are most relevant to the geometric uncertainty. Then We compute the sensitivity score $U_i$ for each surfel as the $L_1$-norm of the log of the diagonal elements of its Fisher information matrix:

\vspace{-1em}
\begin{equation}
U_i = ||\log(\text{diag}( \sum_{\phi \in P_{gt}} \nabla_{x_i,s_i} \textbf{I}_{\mathcal{G}_i} \nabla_{x_i,s_i} \textbf{I}^T_{\mathcal{G}_i})||_1
\end{equation}
\vspace{-1em}

This process enables us to identify less impactful surfels while maintaining rendering quality and view consistency through the ray sorting mechanism.

\subsection{Per-Surfel Texture}
The per-ray sorting strategy functions as a regularization to stabilize the multi-view rendering consistency, which increases pressure in the appearance fitting performance as studied by ~\cite{radl2024stopthepop}. Meanwhile, the adaptive control procedures in Gaussian optimization prunes low opacity primitives which tend to model the high-frequency details.
We assign an optimizable texture map, rather than a single RGB color to each surfel at the second stage to overcome this drawback. The benefit of texture maps can be explained in two domains: 1) Adding more optimizable parameters to balance the performance drop introduced by the per-ray depth sorting regularization. 2) Enabling each surfel to model a more complex and detailed appearance.

Given a surfel with scaling factors $(s_u, s_v)$, we assign it with a texture map $\mathbf{C}[u, v]$ of size $(U, V, c),\; U = \left \lceil Ts_u  \right \rceil, \, V= \left \lceil Ts_v  \right \rceil$, where $c$ is the number of color channels, and $T$ is a hyperparameter determined by the memory and number of primitives empirically. A texture resolution upper bound is set to constraint memory consumption and all textures are initialized by the background color. During rendering, we only use the center ray for each pixel frustum to query the texture map. For each ray-surfel intersection with surfel coordinates $(u, v)$ we set a cutoff range $r$ as 4.5 and compute the indexing coordinate as $(\frac{u+r}{2r}U, \frac{v+r}{2r}V)$ to bilinearly sample the texture map to get the 0-th component of the Spherical Harmonics. Different from ~\cite{Huang2DGS2024, rong2024gstex, kerbl3Dgaussians}, we compute the view direction of the surfel as the precise camera-intersection direction $\mathbf{d} = \mathbf{H}(u, v, 1)^\top - \mathbf{O}$, rather than the approximated camera-center direction $\mathbf{p} - \mathbf{O}$ considering large surfels, where $\mathbf{O}$ is the camera center in world coordinates. Now the per-pixel alpha blending equation is formulated as Eq.\ref{eq:blending}, where the index $i$ follows the k-sorted order, and $\mathbf{SH}$ is the spherical parameter rendering.
\begin{equation}
    \mathbf{\hat{C}(x)} = \mathbf{C}[u, v] + \mathbf{SH}(\mathbf{d})
\end{equation}
\begin{equation}
    \label{eq:blending}
    \mathbf{C(x)} = \sum_{i}\mathbf{\hat{C}(x)}_i \alpha_i \hat{G}_i(\mathbf{x}) T_i
\end{equation}
\vspace{-5mm}

\subsection{Optimization}

Our method contains 2 optimization stages as shown in Fig.~\ref{fig:pipeline}: 1) Train a representation whose primitives have only a single color with Fisher pruning. 2) Assign each surfel with texture, and finetune the representation with on-demand higher-resolution images. We follow the 2D Gaussian Splatting for the training target

\vspace{-1em}
\begin{equation}
    L =  L_c + \lambda_n L_n + \lambda_d L_d
\end{equation}
\vspace{-5mm}

where $L_c$ is the appearance reconstruction loss which sums the $L_1$ loss and the SSIM loss, and $L_n, L_d$ are geometric regularization as Eq.\ref{eq:Ld},\ref{eq:Ln}. During the second stage, we stop the gradient of Gaussian centers, rotations, and scales, as the high-frequency rendering responds to geometry modification significantly.

%% file: tables/multires.tex
\begin{table*}[]
    \renewcommand{\tabcolsep}{1pt}
    \centering
    \resizebox{0.95\linewidth}{!}{
    \begin{tabular}{@{}l@{\,\,}|ccccc|ccccc|ccccc}
    & \multicolumn{5}{c|}{PSNR $\uparrow$} & \multicolumn{5}{c|}{SSIM $\uparrow$} & \multicolumn{5}{c}{LPIPS $\downarrow$}  \\
    & Full Res. & 1/2 Res. & 1/4 Res. & 1/8 Res. & Avg. & Full Res. & 1/2 Res. & 1/4 Res. & 1/8 Res. & Avg. & Full Res. & 1/2 Res. & 1/4 Res. & 1/8 Res & Avg.  \\ \hline

NeRF~\cite{mildenhall2020nerf}&     31.48 & 32.43 &  30.29 &  26.70 & 30.23 & 0.949 & 0.962 & 0.964 &  0.951 & 0.956 & 0.061 & 0.041 & 0.044 & 0.067 & 0.053
\\
MipNeRF~\cite{barron2022mip}& \textbf{33.08} &  \textbf{33.31} &  \textbf{30.91} &  \textbf{27.97} & \textbf{ 31.31} & \textbf{0.961} &  \textbf{0.970} &  \textbf{0.969} &  \textbf{0.961} &  \textbf{0.965} & \textbf{0.045} &  \textbf{0.031} &  \textbf{0.036} &  \textbf{0.052} &  \textbf{0.041} \\

\hline
3DGS~\cite{kerbl3Dgaussians}&  33.33 & 26.95 & 21.38 & 17.69 & 24.84 &  0.969 & 0.949 & 0.875 & 0.766 & 0.890 &  \textbf{0.030} & 0.032 & 0.066 & 0.121 & 0.063\\

MipSplatting ~\cite{yu2024mip}&    \textbf{33.36} &  \textbf{34.00} & \textbf{31.85} & \textbf{28.67} & \textbf{31.97} & 0.969 & \textbf{0.977} & \textbf{0.978} & \textbf{0.973} & \textbf{0.974} &  0.031 & \textbf{0.019} & \textbf{0.019} & \textbf{0.026} & \textbf{0.024} 
\\

\hline

2DGS ~\cite{Huang2DGS2024} & 32.67 & 27.19 & 20.57 &  16.65 &  24.27 & 0.967 &  0.949 &  0.851 &  0.717 &  0.871 & 0.035 & 0.038 &  0.085 & 0.155 & 0.078 \\

Ours & \textbf{33.46} &  \textbf{32.16} & \textbf{28.18} & \textbf{23.98} & \textbf{29.45} & \textbf{0.968} &\textbf{ 0.969} & \textbf{0.951} & \textbf{0.913} & \textbf{0.950} & \textbf{0.030} & \textbf{0.027} &  \textbf{0.049} &  \textbf{0.088} &  \textbf{0.049} \\

    \end{tabular}
    }
    \vspace{-0.1in}
    \caption{
    Reduced resolution rendering on the NeRF synthetic dataset~\cite{mildenhall2020nerf}. All methods are trained on full-resolution images and evaluated at four lower resolutions to simulate zoom-out effects. Our method consistently outperforms 2DGS across different resolutions.
    }
    \label{tab:avg_blender_results_single_train_multi_test}
    \vspace{-1.5em}
\end{table*}

%% file: sec/4_exp.tex
\section{Experiments}
\label{sec:exp}

\input{tables/360nerf}

\begin{figure*}
\vspace{-1em}
    \centering
    \includegraphics[width=1.0\linewidth]{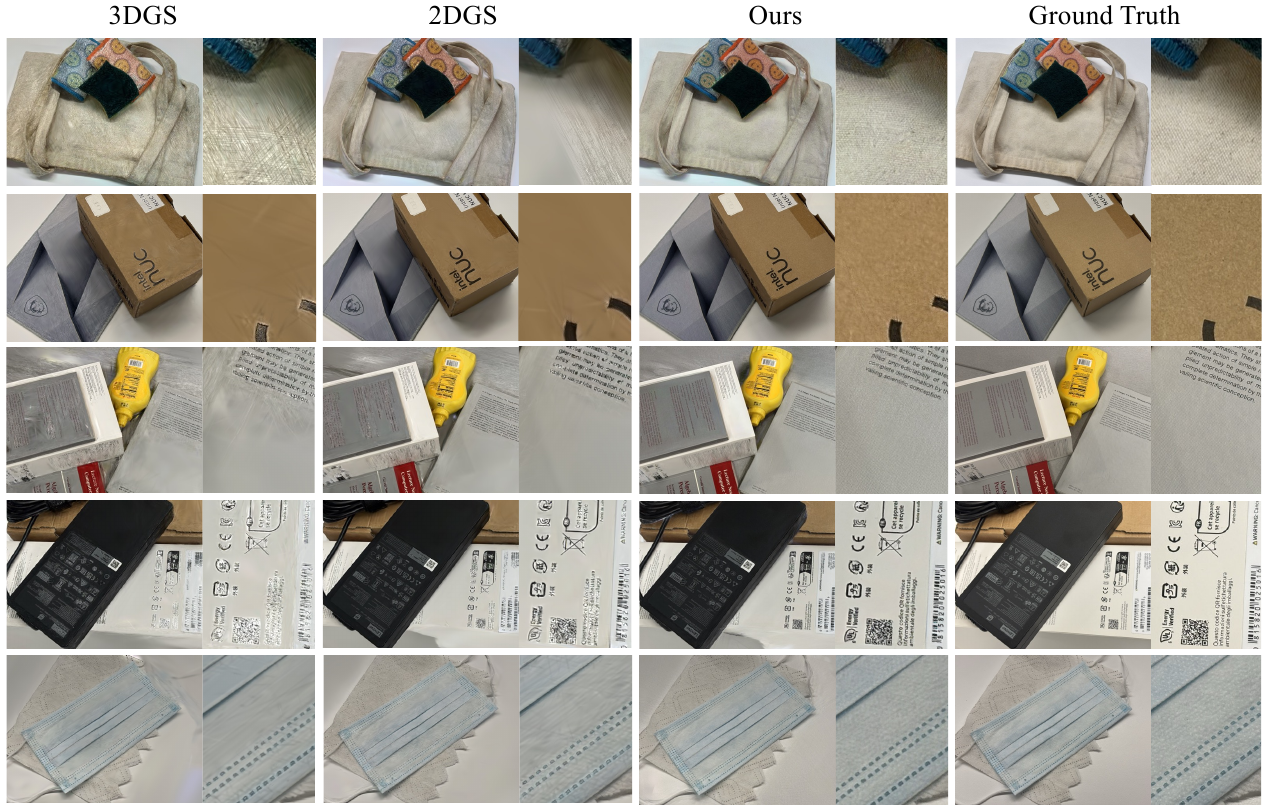}
    \caption{Quanlitative comparison of novel view rendering on five custom scenes featuring materials with intricate details of our method with 3D Gaussian Splatting (3DGS) and 2D Gaussian Splatting (2DGS). Each row showcases a distinct texture-rich material novel view rendering:  1) the fabric texture on a bag, 2) the paper surface of a cardboard box, 3) the text and texture of the book cover, 4) the label details on a box, 5) fine texture of a face mask. The four columns display results from 3DGS, 2DGS, our method, and the ground truth. Across all scenes, our approach consistently captures finer details, demonstrating superior texture fidelity compared to 3DGS and 2DGS.}
    \label{fig:detail}
    \vspace{-1em}
\end{figure*}

\input{tables/custom}

\subsection{Evaluation Setup}
We evaluate our pipeline through comparisons of both appearance and geometry against state-of-the-art methods. We choose the Mip-NeRF 360  ~\cite{barron2022mip} dataset and the NeRF synthetic  \cite{mildenhall2020nerf} dataset for appearance comparison, and DTU dataset~\cite{jensen2014large} for geometry comparison. Additionally, we present results on a self-captured, texture-rich dataset. Unless stated otherwise, we prune 10\% of the surfels after the original 2DGS training. We set the sampling weights $w_m$ and $w_c$ as 0.2, and the texture resolution $T$ as 1000 for the NeRF synthetic dataset, 400 for Mip-NeRF 360 indoor scenes, 200 for outdoor scenes, and 4000 for our texture-rich dataset. We compare our method with 3DGS-based methods like StopPop ~\cite{radl2024stopthepop}, PGSR ~\cite{chen2024pgsr}, and 2DGS-based methods like GSTex~\cite{rong2024gstex}. We evaluate the PSNR, SSIM, and L-PIPS for novel view rendering, and measure the Chamfer distance for geometric reconstruction. We conduct all the experiments on a single NVIDIA L40 GPU.

\input{tables/dtu}

\subsection{Experiment Result}

\begin{figure}
\vspace{-0.5em}
    \centering
    \includegraphics[width=0.8\linewidth]{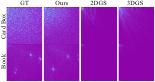}
    \caption{Visualization of DCT on the texture-rich area of our customized dataset. Right-top represents low-frequency, right-bottom represents low-frequency. Our method preserves more high-frequency details than baselines.}
    \label{fig:DCT}
    \vspace{-2em}
\end{figure}

\textbf{Quantitative comparison.} Tab.~\ref{tab:DTU} illustrates that our method sustains a geometric reconstruction quality comparable to the baseline on the DTU dataset ~\cite{jensen2014large}. For appearance reconstruction, we evaluate our approach on the Mip-NeRF 360 ~\cite{barron2022mip} dataset and the NeRF synthetic dataset \cite{mildenhall2020nerf}, with results presented in Tab.~\ref{tab:360}. Our method achieves superior appearance metrics compared to 2DGS while remaining competitive with methods emphasizing appearance over geometry, such as 3DGS \cite{kerbl3Dgaussians} and StopThePop \cite{radl2024stopthepop}. Our method reaches a mean primitive number of $1.3\times10^5$ on the NeRF synthetic dataset and $2.6\times10^6$ on the Mip-NeRF 360 dataset, compared with 2DGS which has $1.4\times10^5$ and $2.2\times10^6$ and MipSplatting which has $3.0\times10^5$ and $4.2\times10^6$ respectively.

To further analyze the high-frequency texture expressivity of our method, we apply the Discrete Cosine Transform (DCT) to texture-rich regions of both rendered and ground truth images.  By measuring the cosine distance of the non-constant components, we quantify high-frequency fidelity as defined in Eq.\ref{eq:hifi}, where $\mathbf{\hat{I}}$ denotes the rendered image and $\mathbf{I}$ is the ground truth image. The results shown in Tab.~\ref{tab:custom} indicate our method preserves more high-frequency details than 2DGS and 3DGS.
\vspace{-0em}
\begin{equation}
    \label{eq:hifi}
    ||HF|| = \frac{DCT_{AC}(\mathbf{\hat{I}})\cdot DCT_{AC}(\mathbf{I})}{||DCT_{AC}(\mathbf{\hat{I}})||_2 ||DCT_{AC}(\mathbf{I})||_2}
\end{equation}

\begin{figure*}
\vspace{-1em}
    \centering
    \includegraphics[width=0.9\linewidth]{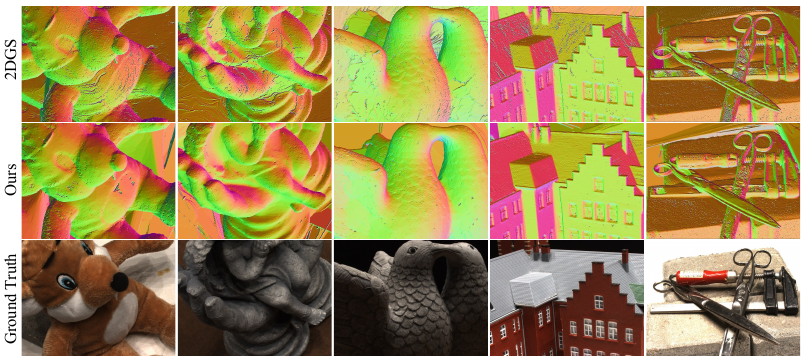}
    \caption{Surface normal comparison between our method, 2DGS, and ground truth appearance in \textbf{novel views} on DTU ~\cite{jensen2014large} dataset. We split the test and training set as ~\cite{li2023neuralangelo}. Our approach achieves smoother surfaces and enhances geometric consistency relative to 2DGS.}
    \label{fig:normal}
    
\end{figure*}

\noindent\textbf{Qualitative Comparison.} Fig.~\ref{fig:alias} illustrates the rendering quality at reduced resolutions (1/2, 1/4 and 1/8), where our method significantly reduces the aliasing artifacts commonly observed in 2DGS. We show the result on the NeRF synthetic dataset as the scene includes rich details on the surface. Tab.~\ref{fig:normal} presents a comparison of novel view surface normals computed from depth gradients between our method and 2DGS, demonstrating that our approach achieves smoother surfaces and enhances geometric consistency. Fig.~\ref{fig:detail} showcases various texture-rich scenes from our custom dataset, including the fabric of a bag, a cardboard texture, a detailed book cover, a labeled box, and the fine texture of a face mask. For these challenging materials, our method consistently outperforms both 3DGS and 2DGS, effectively preserving high-frequency details. The DCT result of the cardboard box and the book is shown in Fig.~\ref{fig:DCT}, highlighting that our method retains more high-frequency details than the competing methods.

\subsection{Ablation Study}

We conduct an ablation study to evaluate the impact of key design choices on appearance reconstruction using the NeRF synthetic dataset \cite{mildenhall2020nerf}. We examine four crucial components: texture maps, Fisher-based pruning, ray sorting, and frustum-based anti-aliasing. Tab.~\ref{tab:ablation} presents the quantitative results when removing each component. Fig.~\ref{fig:ablation} shows the qualitative results when removing frustum-based sampling, texture, and ray sorting. We analyze the impact of removing each one of the components:

\begin{figure}
\vspace{-1em}
    \centering
    \includegraphics[width=0.9\linewidth]{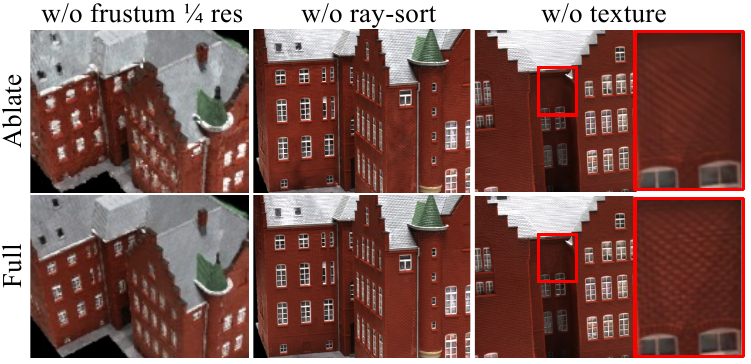}
    \caption{Qualitative ablation study. We show the qualitative result on the test views of scan24 in the DTU dataset ~\cite{jensen2014large} when removing frustum-based sampling, texture, and ray sorting, respectively.}
    \label{fig:ablation} 
    \vspace{-2em}
\end{figure}

\textbf{Texture Maps:} Removing texture maps results in decreased PSNR and increased LPIPS, with particularly noticeable degradation in scenes containing fine details. This validates our hypothesis that per-primitive texture maps enable better modeling of high-frequency appearance details.

\textbf{Fisher information-based Pruning:} 
~\cite{radl2024stopthepop} introduces an approach to calculating the sorting error of a rendering. For each ray, it computes a per-ray depth order error $\epsilon$ as:

\vspace{-0.5em}
\begin{equation} \label{eq:sorterr}
\epsilon = \sum_{i} (z_{i} - z_{i+1}) \cdot \mathbbm{1}[z_{i+1} < z_{i}]
\end{equation}
\vspace{-1em}

where $z_{i}$ is the blending depth of the $i$-th surfel along the ray, and $\mathbbm{1}[\cdot]$ is the indicator function when consecutive surfels are out of order. The error is averaged over all the rays on one image. Our analysis demonstrates that incorporating Fisher information pruning reduces the sorting order error for both training and testing datasets, as shown in Tab.~\ref{tab:sorterr}.

\textbf{Ray Sorting:} Ray sorting proves crucial for our approach, as its removal causes the most significant drop across all metrics, as the popping artifacts become particularly severe when the scene is overfitted by per-surfel texture without sorting.

\textbf{Frustum-based Sampling:} While removing the frustum-based sampling strategy only slightly impacts the metrics, we find this design particularly valuable for low-resolution and distant view rendering as shown in Fig.~\ref{fig:ablation}.

\input{tables/ablation}

%% file: tables/360nerf.tex
\begin{table}[t]
    \label{360nerf}
    \centering
    \small
    \resizebox{\columnwidth}{!}{%
    \begin{tabular}{l|ccc|ccc}
        \toprule
        Method & \multicolumn{3}{c|}{Mip-NeRF 360~\cite{barron2022mip}} & \multicolumn{3}{c}{NeRF synthetic~\cite{mildenhall2020nerf} } \\
         & PSNR \(\uparrow\) & SSIM\(\uparrow\) & LPIPS \(\downarrow\) & PSNR \(\uparrow\) & SSIM\(\uparrow\) & LPIPS \(\downarrow\) \\
        \midrule
        3DGS \cite{kerbl3Dgaussians} & 27.24 & 0.815 & 0.214 & 33.33 & 0.969 & 0.030 \\
        StopPop \cite{radl2024stopthepop}&27.27& 0.814 & 0.213 & \textbf{33.57} & \textbf{0.970} & \textbf{0.030} \\
        PGSR \cite{chen2024pgsr}&  27.43 & \textbf{0.830} & 0.193 & 31.87 & 0.964 & 0.035 \\
        MipSplatting\cite{yu2024mip} & \textbf{27.73} & 0.828 & \textbf{0.189}    & 33.36 &   0.969   & 0.031 \\
        \midrule
        2DGS \cite{Huang2DGS2024} & 27.03 & 0.805 & 0.223 & 32.67 & 0.967 & 0.035 \\
        GSTex \cite{rong2024gstex} & -   & - & - & 33.25 & \textbf{0.969} & \textbf{0.024} \\
        Ours & \textbf{27.25} & \textbf{0.807} & \textbf{0.216} & \textbf{33.46} & 0.968 & 0.030\\ 
        \bottomrule
    \end{tabular}
    }
    \vspace{-1em}
    \caption{Mean quantitative results on Mip-NeRF 360~\cite{barron2022mip} and NeRF sythetic~\cite{mildenhall2020nerf} dataset.}
    \label{tab:360}
    \vspace{-0.5em}
\end{table}

%% file: tables/custom.tex
\begin{table}[t]
    \centering
    \small
    \begin{tabular}{l|cccc}
        \toprule
        Method & PSNR \(\uparrow\) & SSIM\(\uparrow\) & LPIPS \(\downarrow\) & $||HF||$ \(\uparrow\) \\
        \midrule
        3DGS \cite{kerbl3Dgaussians} & 22.40 & 0.602 & 0.461 & 0.304 \\
        2DGS \cite{Huang2DGS2024} & 22.11 & \textbf{0.627} & 0.460  & 0.342\\
        Ours & \textbf{22.72} & 0.607 & \textbf{0.350} & \textbf{0.434}\\
        \bottomrule
    \end{tabular}
    \vspace{-1em}
    \caption{Quantitative results on our texture-rich dataset.}
    \label{tab:custom}
    \vspace{-0.5em}
\end{table}

%% file: tables/dtu.tex
\begin{table}
    \centering
    \small
    \begin{minipage}{0.57\columnwidth}
        \centering
        \begin{tabular}{c|cc}
            \toprule
             order err $\epsilon$ & Train & Test \\
            \midrule
            2DGS \cite{Huang2DGS2024} & 0.2392   &   0.2713 \\
            Ours w/o prune & 0.0023 & 0.0024 \\
            Ours w/ prune & 0.0021 & 0.0021 \\
            \bottomrule
        \end{tabular}
        \vspace{-1em}
        \caption{The train and test views average sorting error after k-sorting of the NeRF synthetic dataset.}
        \label{tab:sorterr}
    \end{minipage}
    \hfill
    \begin{minipage}{0.4\columnwidth}
        \centering
        \begin{tabular}{l|c}
            \toprule
            Method & CD \\
            \midrule
            2DGS \cite{Huang2DGS2024} & 0.74 \\
            Ours & 0.75 \\
            \bottomrule
        \end{tabular}
        \caption{The Chamfer distance measured on the DTU~\cite{jensen2014large} dataset.}
        \label{tab:DTU}
    \end{minipage}%
    \vspace{-1em}
\end{table}

%% file: tables/ablation.tex
\begin{table}[t]
\label{nerf}
    \vspace{-1em}
    \centering
    \scriptsize
    \begin{tabular}{l|ccc}
        \toprule
        Method & PSNR \(\uparrow\) & SSIM \(\uparrow\) & LPIPS \(\downarrow\)  \\
        \midrule
        full & 33.46 & 0.968 & 0.030 \\
        w/o texture & 33.28 & 0.968 & 0.032 \\
        w/o prune & 33.40 & 0.967 & 0.031 \\
        w/o sorting & 31.99 & 0.955 & 0.048 \\
        w/o frustum & 33.27 & 0.967 & 0.031\\
        \bottomrule
    \end{tabular}
    \vspace{-1em}
    \caption{Ablation study on NeRF synthetic  \cite{mildenhall2020nerf} dataset.}
    \label{tab:ablation}
    \vspace{-2em}
\end{table}

%% file: sec/5_conclusion.tex
\section{Conclusions}
\label{sec:conclusion}
In conclusion, our proposed pipeline achieves high-fidelity scene rendering by addressing limitations in detailed texture preservation and anti-aliasing that have challenged previous Gaussian splatting approaches. Through frustum-based sampling, Fisher-based pruning, per-ray sorting, and per-surfel texture mapping, our technique achieves robust performance across benchmark datasets and our texture-rich scenes.

\textit{Limitation}. Despite the strong performance of our method in preserving fine details and reducing artifacts, the computational cost of our per-ray sorting and texture map indexing remains higher than baseline methods, which could impact scalability in real-time applications or on devices with limited GPU resources. 

\textbf{Acknowledgement}
The authors appreciate the support of the gift from AWS AI to Penn Engineering's ASSET Center for Trustworthy AI; 
and the support of  the following grants: 
NSF IIS-RI 2212433, NSF FRR 2220868 awarded to UPenn.

%% file: sec/X_suppl.tex
\clearpage
\setcounter{page}{1}
\maketitlesupplementary

\section{Further ablation}

We present a comprehensive ablation study in Tab.~\ref{tab:ablation_full} to evaluate the impact of key design choices on appearance reconstruction. Additionally, we report results from another ablation study conducted on our customized texture-rich dataset, examining the effect of removing per-surfel texture in Tab.~\ref{tab:ablation_custom}. The influence of different pruning percentages is analyzed in Tab.~\ref{tab:ablation_prune}.

\input{tables/custom_supp}

\input{tables/prune_full}

\section{Sampling Strategy Error Analysis}

\begin{figure}
    \centering
    \includegraphics[width=0.9\linewidth]{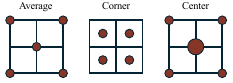}
    \caption{Visualization of different sampling strategy}
    \label{fig:sample}
    \vspace{-1em}
\end{figure}

There are multiple alternative sampling strategies to numerically approximate the average density, e.g. the integral value of the intersection of the pixel frustum and the 2D Gaussian surfel. Given the sampling points in a pixel $P$, e.g. $(x_i, y_i) \in [-0.5, 0.5]^2$, we can determine the parameters $w_i$ to approximate the integrated value by the polynomials of the sampled function value on sampling points as Eq.~\ref{eq:approx}. 
\begin{equation}
    \iint_P f(x, y)dxdy = \sum_iw_if(x_i, y_i) + O((x+y)^n)
    \label{eq:approx}
\end{equation}

Through the Taylor Series, we have the estimated integral function value $f$ in a small region $P = h\times h$ around the origin $(0, 0)$ as Eq.~\ref{eq:inte}.
\begin{equation}
\begin{aligned}
        & \int_{-h/2}^{h/2}\int_{-h/2}^{h/2}f(x, y)dxdy \\ 
        =& \iint_P\left(f(0, 0)+f'_x(0, 0)x+f'_y(0, 0)y+O((x+y)^2)\right)dxdy \\
        =& h^2f(0, 0)+ \frac{h^4f''_{xx}(0, 0)}{24}+\frac{h^4f''_{yy}(0, 0)}{24}+O(h^6)
\end{aligned}
\label{eq:inte}
\end{equation}

As the function value at $(x_i, y_i) \in \{\pm h/2, \pm h/2\}$ can be written as the approximation of the Taylor Expansion at $(0, 0)$, we can analyze and cancel the higher order errors by the linear combination of sampled values. For 4 corner points, we have 
\begin{equation}
\begin{aligned}
        & \frac{h^2}{4}\left(f(\frac{-h}{2}, \frac{-h}{2})+f(\frac{h}{2}, \frac{-h}{2})+f(\frac{-h}{2}, \frac{h}{2})+f(\frac{h}{2}, \frac{h}{2})\right)\\
        = & h^2f(0, 0)+\frac{h^4f''_{xx}(0, 0)}{8}+\frac{h^4f''_{yy}(0, 0)}{8} + O(h^6) \\
        = & \iint_P f(x, y)dxdy + \frac{h^4f''_{xx}(0, 0)}{12}+\frac{h^4f''_{yy}(0, 0)}{12} + O(h^6)\\ 
        = & \iint_P f(x, y)dxdy + O(h^4)
\end{aligned}
\label{eq:quarter}
\end{equation}

\begin{figure*}
    \centering
    \includegraphics[width=0.9\linewidth]{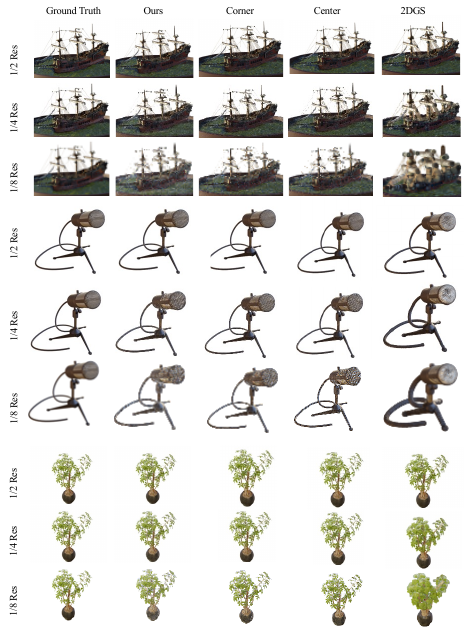}
    \caption{Reduced resolution rendering on the NeRF synthetic dataset~\cite{mildenhall2020nerf} for different sampling strategy: Ours(average), Corner(Eq.~\ref{eq:quarter}), Center(Eq.~\ref{eq:812}). All methods are trained on full-resolution images and evaluated at four lower resolutions to simulate zoom-out effects.}
    \label{fig:sampling_stg_alias}
\end{figure*}

This means all the $w_i$ equals $\frac{1}{4}$ in Eq.~\ref{eq:approx} as illustrated in Fig.~\ref{fig:sample} the Corner case. To further cancel the 4-th order error, we can involve the function value at $(0, 0)$. The error term of $f(0, 0)$ can be written as 
\begin{equation}
\begin{aligned}
h^2f(0, 0) = & \iint_P f(x, y)dxdy - \frac{h^4f''_{xx}(0, 0)}{24} -\frac{h^4f''_{yy}(0, 0)}{24} \\
            & + O(h^6)\\
\end{aligned}
\end{equation}
So we can cancel the $o(h^4)$ error item by 
\begin{equation}
\begin{aligned}
    & \frac{h^2}{12}\left(8f(0, 0) + \sum_{i, j\in\{-1, 1\}}f(i\frac{h}{2}, j\frac{h}{2})\right)  \\
    = & \iint_P f(x, y)dxdy + O(h^6)
\end{aligned}
    \label{eq:812}
\end{equation}
This means the weights in Eq.~\ref{eq:approx} are $\frac{1}{12}$ and $\frac{2}{3}$ for the four corner points and the center point respectively, as illustrated in Fig.~\ref{fig:sample} the Center case. However, the theoretical error may not always hold in real scenarios, where the vanilla average of 5 points used in the main paper may be more robust. We report the rendering quality of different strategies in different resolutions in Tab.~\ref{tab:sample} and show the qualitative results on selected scenes in Fig.~\ref{fig:sampling_stg_alias}. The strategy we adopt in the main paper performs best on the full resolution, while other methods have better results on some other resolutions. As a result, different strategies can be adopted for various tasks.

\input{tables/multires_supp}

\section{Full results}

We show the full comparison of our models versus previous works on the DTU dataset ~\cite{jensen2014large}, Mip-NeRF 360 ~\cite{barron2022mip} dataset and the NeRF synthetic dataset \cite{mildenhall2020nerf} in Tab.~\ref{tab:dtu_full},Tab.~\ref{tab:360_full}, Tab.~\ref{tab:nerf_full}, respectively.

\input{tables/360_full}
\input{tables/dtu_full}
\input{tables/nerf_full}
\input{tables/ablation_full}

%% file: tables/custom_supp.tex
\begin{table}[t]
    \centering
    \small
    \begin{tabular}{l|cccc}
        \toprule
        Method & PSNR \(\uparrow\) & SSIM\(\uparrow\) & LPIPS \(\downarrow\) & $||HF||$ \(\uparrow\) \\
        \midrule
        3DGS \cite{kerbl3Dgaussians} & 22.40 & 0.602 & 0.461 & 0.304 \\
        2DGS \cite{Huang2DGS2024} & 22.11 & \textbf{0.627} & 0.460  & 0.342\\
        Ours & \textbf{22.72} & 0.607 & \textbf{0.350} & \textbf{0.434}\\
        Ours w/o texture & 22.58 & 0.624 & 0.455 & 0.345 \\
        \bottomrule
    \end{tabular}
    \caption{Quantitative results on our texture-rich dataset.}
    \label{tab:ablation_custom}
    \vspace{-1em}
\end{table}

%% file: tables/prune_full.tex
\begin{table*}[t]
\centering
\begin{tabular}{l|cccccccc|c}
\toprule
Pruned \% & chair & drums & ficus & hotdog & lego & materials & mic & ship & mean \\
\midrule
10\% (Ours) & 35.92 & 26.05 & 35.90 & 38.08 & 35.37 & 29.87 & 35.36 & 31.10 & \textbf{33.46} \\
20\% & 35.90 & 26.06 & 35.91 & 38.04 & 35.31 & 29.85 & 35.33 & 31.09 & 33.44 \\
30\% & 35.89 & 26.05 & 35.83 & 38.02 & 35.39 & 29.83 & 35.37 & 31.12 & 33.44 \\
40\% & 35.83 & 26.00 & 35.80 & 37.84 & 35.34 & 29.71 & 35.36 & 31.14 & 33.38 \\
50\% & 35.69 & 25.87 & 35.73 & 37.52 & 35.16 & 29.45 & 35.34 & 31.19 & 33.24 \\
\midrule
10\%  (Ours) & 0.986 & 0.952 & 0.988 & 0.986 & 0.981 & 0.957 & 0.991 & 0.901 & \textbf{0.968} \\
20\% & 0.986 & 0.952 & 0.988 & 0.986 & 0.981 & 0.957 & 0.991 & 0.901 & 0.968 \\
30\% & 0.986 & 0.953 & 0.988 & 0.986 & 0.981 & 0.956 & 0.991 & 0.902 & 0.968 \\
40\% & 0.986 & 0.952 & 0.988 & 0.986 & 0.980 & 0.954 & 0.991 & 0.902 & 0.967 \\
50\% & 0.985 & 0.951 & 0.988 & 0.986 & 0.979 & 0.951 & 0.991 & 0.903 & 0.967 \\

\midrule
10\% (Ours) & 0.013 & 0.042 & 0.011 & 0.019 & 0.017 & 0.037 & 0.007 & 0.094 & \textbf{0.030} \\
20\% & 0.013 & 0.042 & 0.011 & 0.019 & 0.017 & 0.037 & 0.007 & 0.094 & 0.030 \\
30\% & 0.013 & 0.042 & 0.011 & 0.018 & 0.017 & 0.038 & 0.007 & 0.094 & 0.030 \\
40\% & 0.014 & 0.043 & 0.011 & 0.018 & 0.018 & 0.040 & 0.007 & 0.095 & 0.031 \\
50\% & 0.015 & 0.045 & 0.012 & 0.019 & 0.019 & 0.043 & 0.007 & 0.096 & 0.032 \\
\bottomrule
\end{tabular}
\caption{Pruning ablation on NeRF synthetic dataset \cite{mildenhall2020nerf}, We report PSNR \(\uparrow\), SSIM \(\uparrow\), LPIPS \(\downarrow\) respectively.}
\label{tab:ablation_prune}
\end{table*}

%% file: tables/multires_supp.tex
\begin{table*}[]
    \renewcommand{\tabcolsep}{1pt}
    \centering
    \resizebox{0.95\linewidth}{!}{
    \begin{tabular}{@{}l@{\,\,}|ccccc|ccccc|ccccc}
    & \multicolumn{5}{c|}{PSNR $\uparrow$} & \multicolumn{5}{c|}{SSIM $\uparrow$} & \multicolumn{5}{c}{LPIPS $\downarrow$}  \\
    & Full Res. & 1/2 Res. & 1/4 Res. & 1/8 Res. & Avg. & Full Res. & 1/2 Res. & 1/4 Res. & 1/8 Res. & Avg. & Full Res. & 1/2 Res. & 1/4 Res. & 1/8 Res & Avg.  \\ \hline

\hline
3DGS~\cite{kerbl3Dgaussians}&  33.33 & 26.95 & 21.38 & 17.69 & 24.84 &  0.969 & 0.949 & 0.875 & 0.766 & 0.890 &  \textbf{0.030} & 0.032 & 0.066 & 0.121 & 0.063\\

MipSplatting ~\cite{yu2024mip}&    \textbf{33.36} &  \textbf{34.00} & \textbf{31.85} & \textbf{28.67} & \textbf{31.97} & 0.969 & \textbf{0.977} & \textbf{0.978} & \textbf{0.973} & \textbf{0.974} &  0.031 & \textbf{0.019} & \textbf{0.019} & \textbf{0.026} & \textbf{0.024} 
\\

\hline

2DGS ~\cite{Huang2DGS2024} & 32.67 & 27.19 & 20.57 &  16.65 &  24.27 & 0.967 &  0.949 &  0.851 &  0.717 &  0.871 & 0.035 & 0.038 &  0.085 & 0.155 & 0.078 \\

Ours & \textbf{33.46} &  32.16 & 28.18 & 23.98 & 29.45 & \textbf{0.968} & 0.969 & 0.951 & 0.913 & 0.950 & \textbf{0.030} & 0.027 &  0.049 &  0.088 &  0.049 \\

Ours Center Eq.~\ref{eq:812} & 33.26 & 33.29 & \textbf{30.49} & \textbf{26.77} & \textbf{30.95} & 0.967 & 0.972 & 0.964 & 0.943 & 0.962 & 0.030 & \textbf{0.024} & \textbf{0.039} & \textbf{0.073} & \textbf{0.042} \\

Ours Corner Eq.~\ref{eq:quarter} & 33.35 &  \textbf{33.35} & 30.45 & 26.61 & 30.94 & 0.967 & \textbf{0.973} & \textbf{0.966} &  \textbf{0.944} & \textbf{0.963} & 0.030 & 0.025 & 0.042 &  0.075 & 0.043 \\
    \end{tabular}
    }
    \vspace{-0.1in}
    \caption{
    Reduced resolution rendering on the NeRF synthetic dataset~\cite{mildenhall2020nerf}. All methods are trained on full-resolution images and evaluated at four lower resolutions to simulate zoom-out effects. Our method consistently outperforms 2DGS across different resolutions.
    }
    \label{tab:sample}
    
\end{table*}

%% file: tables/360_full.tex
\begin{table*}[th!]
    \centering
    \begin{tabular}{l|ccccc|cccc|c}
        \toprule
        & bicycle & flowers & garden & stump & treehill & room & counter & kitchen & bonsai & mean \\
        \midrule
        Ours & 24.83 & 21.22 & 26.98 & 26.40 & 22.50 & 31.20 & 28.94 & 31.09 & 32.06 & 27.25 \\
        2DGS \cite{Huang2DGS2024}& 24.82 & 20.99 & 26.91 & 26.41 & 22.52 & 30.86 & 28.45 & 30.62 & 31.64 & 27.03 \\
        3DGS \cite{kerbl3Dgaussians} & 24.71 & 21.09 & 26.63 & 26.45 & 22.33 & 31.50 & 29.07 & 31.13 & 32.26 & 27.24 \\
        StopPop \cite{radl2024stopthepop}&25.20 & 21.50 & 27.16 & 26.69 & 22.43 & 30.83 & 28.59 & 31.13 & 31.93 & \textbf{27.27} \\
        \midrule
        Ours & 0.742 & 0.588 & 0.857 & 0.760 & 0.608 & 0.925 & 0.909 & 0.928 & 0.943 & 0.807 \\
        2DGS \cite{Huang2DGS2024}& 0.752 & 0.588 & 0.852 & 0.765 & 0.627 & 0.912 & 0.900 & 0.919 & 0.933 & 0.805 \\
        3DGS \cite{kerbl3Dgaussians} & 0.771 & 0.605 & 0.868 & 0.775 & 0.638 & 0.914 & 0.905 & 0.922 & 0.938 & \textbf{0.815} \\
        StopPop&0.767& 0.604 & 0.862 & 0.775 & 0.635 & 0.917 & 0.903 & 0.925 & 0.939 & 0.814 \\
        \midrule
        Ours & 0.231 & 0.340 & 0.115 & 0.223 & 0.354 & 0.201 & 0.188 & 0.119 & 0.177 & 0.216 \\
        2DGS \cite{Huang2DGS2024}& 0.218 & 0.346 & 0.115 & 0.222 & 0.329 & 0.223 & 0.208 & 0.133 & 0.214 & 0.223 \\
        3DGS \cite{kerbl3Dgaussians} & 0.205 & 0.336 & 0.103 & 0.210 & 0.317 & 0.220 & 0.204 & 0.129 & 0.205 & 0.214 \\
        StopPop \cite{radl2024stopthepop}&0.206& 0.335 & 0.107 & 0.210 & 0.319 & 0.216 & 0.200 & 0.126 & 0.202 & \textbf{0.213} \\
        \bottomrule
    \end{tabular}
    \caption{Quantitative results on Mip-NeRF 360~\cite{barron2022mip} dataset. We report PSNR \(\uparrow\), SSIM \(\uparrow\), LPIPS \(\downarrow\), respectively.}
    \label{tab:360_full}
\end{table*}

%% file: tables/dtu_full.tex
\begin{table*}[t]
    \centering
    \begin{tabular}{l|ccccccccccccccc|c}
        \toprule
        Scene & 24 & 37 & 40 & 55 & 63 & 65 & 69 & 83 & 97 & 105 & 106 & 110 & 114 & 118 & 122 & Mean \\
        \midrule
        Ours & 0.45 & 0.80 & 0.33 & 0.37 & 0.89 & 0.91 & 0.76 & 1.32 & 1.23 & 0.67 & 0.64 & 1.19 & 0.38 & 0.71 & 0.56 & 0.75 \\
        2DGS \cite{Huang2DGS2024} & 0.46 & 0.80 & 0.33 & 0.37 & 0.95 & 0.86 & 0.80 & 1.25 & 1.24 & 0.67 & 0.67 & 1.24 & 0.39 & 0.64 & 0.47 & \textbf{0.74} \\
        \bottomrule
    \end{tabular}
     \caption{Quantitative result on the DTU Dataset~\cite{jensen2014large} on Chamfer distance \(\downarrow\).}
     \label{tab:dtu_full}
\end{table*}

%% file: tables/nerf_full.tex
\begin{table*}[t]
\centering
\begin{tabular}{l|cccccccc|c}
\toprule
Method & chair & drums & ficus & hotdog & lego & materials & mic & ship & mean \\
\midrule
3DGS \cite{kerbl3Dgaussians}& 35.90 & 26.16 & 34.85 & 37.70 & 35.78 & 30.00 & 35.42 & 30.90 & 33.34 \\
StopPop \cite{radl2024stopthepop}& 35.70 & 26.23 & 35.58 & 37.93 & 36.30 & 30.50 & 36.53 & 31.77 & \textbf{33.57}\\
PGSR \cite{chen2024pgsr} &  34.90 & 25.83 & 31.99 & 35.92 & 33.74 & 28.80 & 34.17 & 29.64 & 31.87 \\
GSTex \cite{rong2024gstex} & 35.51 & 26.06 & 35.65 & 37.49 & 35.51 & 29.72 & 35.28 & 30.80 & 33.25 \\
2DGS & 34.24 & 26.10 & 35.40 & 36.97 & 34.03 & 29.54 & 34.64 & 30.46 & 32.67 \\
Ours & 35.92 & 26.05 & 35.90 & 38.08 & 35.37 & 29.87 & 35.36 & 31.10 & 33.46 \\
\midrule
3DGS \cite{kerbl3Dgaussians}& 0.987 & 0.955 & 0.987 & 0.985 & 0.983 & 0.960 & 0.992 & 0.907 & 0.969 \\
StopPop \cite{radl2024stopthepop}& 0.988 & 0.955 & 0.987 & 0.986 & 0.983 & 0.961 & 0.992 & 0.906 & 0.970\\
PGSR \cite{chen2024pgsr} & 0.985 & 0.951 & 0.978 & 0.984 & 0.978 & 0.954 & 0.990 & 0.890 & 0.964 \\
2DGS \cite{Huang2DGS2024} & 0.984 & 0.954 & 0.988 & 0.984 & 0.978 & 0.957 & 0.990 & 0.904 & 0.967\\
GSTex \cite{rong2024gstex} &0.987 & 0.954 & 0.988 & 0.985 & 0.982 & 0.958 & 0.991 & 0.904 & \textbf{0.969 }\\
Ours & 0.986 & 0.952 & 0.988 & 0.986 & 0.981 & 0.957 & 0.991 & 0.901 & 0.968 \\
\midrule
3DGS \cite{kerbl3Dgaussians}& 0.012 & 0.037 & 0.012 & 0.020 & 0.016 & 0.034 & 0.006 & 0.107 & 0.030 \\
StopPop \cite{radl2024stopthepop}&0.010 & 0.036 & 0.012 & 0.019 & 0.016 & 0.036 & 0.006 & 0.104 & 0.030\\
PGSR \cite{chen2024pgsr} & 0.015 & 0.039 & 0.020 & 0.022 & 0.020 & 0.042 & 0.008 & 0.116 & 0.035 \\
2DGS \cite{Huang2DGS2024} & 0.017 & 0.040 & 0.012 & 0.024 & 0.023 & 0.039 & 0.008 & 0.114 & 0.035\\
GSTex \cite{rong2024gstex} &0.009 & 0.038 & 0.008 & 0.013 & 0.011 & 0.020 & 0.006 & 0.085 & 0.024 \\
Ours & 0.013 & 0.042 & 0.011 & 0.019 & 0.017 & 0.037 & 0.007 & 0.094 & \textbf{0.030} \\
\bottomrule
\end{tabular}
\caption{Quantitative result on NeRF synthetic dataset \cite{mildenhall2020nerf}, We report PSNR \(\uparrow\), SSIM \(\uparrow\), LPIPS \(\downarrow\) respectively.}
\label{tab:nerf_full}
\end{table*}

%% file: tables/ablation_full.tex
\begin{table*}[t]
\centering
\begin{tabular}{l|cccccccc|c}
\toprule
Method & chair & drums & ficus & hotdog & lego & materials & mic & ship & mean \\
\midrule
Ours & 35.92 & 26.05 & 35.90 & 38.08 & 35.37 & 29.87 & 35.36 & 31.10 & \textbf{33.46} \\
w/o texture & 35.59&26.04&35.81&37.74&35.25&29.86&35.24&30.73&33.28 \\
w/o pruning & 35.80 & 26.06 & 35.96 & 37.95 & 35.30 & 29.82 & 35.32 & 31.02 & 33.40 \\
w/o sorting & 34.16 & 25.63 & 35.12 & 35.76 & 33.23 & 29.05 & 33.34 & 29.64 & 31.99 \\
w/o frustum & 35.66 & 26.06 & 35.89 & 37.23 & 34.99 & 29.75 & 35.43 & 31.19 & 33.27 \\
\midrule
Ours & 0.986 & 0.952 & 0.988 & 0.986 & 0.981 & 0.957 & 0.991 & 0.901 & \textbf{0.968} \\
w/o texture & 0.985 & 0.953 & 0.988 & 0.985 & 0.980 & 0.958 & 0.991 & 0.904 & 0.968   \\
w/o pruning & 0.986 & 0.952 & 0.989 & 0.985 & 0.980 & 0.956 & 0.991 & 0.900 & 0.967   \\
w/o sorting & 0.978 & 0.942 & 0.985 & 0.973 & 0.965 & 0.941 & 0.980 & 0.872 & 0.955   \\
w/o frustum & 0.986 & 0.954 & 0.989 & 0.985 & 0.979 & 0.956 & 0.991 & 0.902 & 0.967   \\
\midrule
Ours & 0.013 & 0.042 & 0.011 & 0.019 & 0.017 & 0.037 & 0.007 & 0.094 & \textbf{0.030} \\
w/o texture & 0.015 & 0.041 & 0.011 & 0.022 & 0.019 & 0.036 & 0.007 & 0.103 & 0.032   \\
w/o pruning & 0.014 & 0.042 & 0.011 & 0.019 & 0.018 & 0.038 & 0.008 & 0.095 & 0.031   \\
w/o sorting & 0.023 & 0.055 & 0.016 & 0.046 & 0.036 & 0.061 & 0.027 & 0.122 & 0.048   \\
w/o frustum & 0.015 & 0.040 & 0.011 & 0.020 & 0.020 & 0.038 & 0.008 & 0.095 & 0.031   \\
\bottomrule
\end{tabular}
\caption{Ablation study on NeRF synthetic dataset \cite{mildenhall2020nerf}, We report PSNR \(\uparrow\), SSIM \(\uparrow\), LPIPS \(\downarrow\) respectively.}
\label{tab:ablation_full}
\end{table*}